\documentclass{article}
\usepackage{ijcai25}

\usepackage{times}
\usepackage{soul}
\usepackage{url}
\usepackage[hidelinks]{hyperref}
\usepackage[utf8]{inputenc}
\usepackage[small]{caption}
\usepackage{graphicx}
\usepackage{amsmath}
\usepackage{amsthm}
\usepackage{booktabs}
\usepackage{algorithm}
\usepackage{algorithmic}
\usepackage[switch]{lineno}
\usepackage{xcolor}
\usepackage{xspace}
\usepackage{colortbl}
\usepackage{pifont}
\usepackage{amsfonts}
\usepackage{xspace}
\usepackage{threeparttable}
\usepackage{enumitem}
\usepackage{multirow}
\usepackage{courier}
\usepackage{subcaption}

\newtheorem{definition}{Definition}

\definecolor{Red}{RGB}{185, 5, 14}
\definecolor{Green}{RGB}{19,175,85}
\definecolor{Blue}{RGB}{46, 96, 179}
\newcommand{\method}[0]{\textbf{\texttt{\textcolor{Red}{T}\textcolor{Green}{2}\textcolor{Blue}{S}}}\xspace}
\newcommand{\dataset}[0]{\texttt{TSFragment-600K}\xspace}
\newcommand{\model}[0]{\textbf{\texttt{\textcolor{Red}{T}\textcolor{Green}{2}\textcolor{Blue}{S}}}\xspace}

\definecolor{tabhighlight}{HTML}{e5e5e5}
\definecolor{someorange}{rgb}{0.36,0.49,0.85}
\definecolor{someblue}{rgb}{0.27, 0.35, 0.760}
\definecolor{somepurple}{rgb}{0.32, 0.42, 0.80}
\let\oldcite\cite
\renewcommand{\cite}[1]{\textcolor{someorange}{\oldcite{#1}}}

\title{\method: High-resolution Time Series Generation with Text-to-Series Diffusion Models}

\author{
Yunfeng Ge$^{1, 2}$, Jiawei Li$^{2, 7}$, Yiji Zhao$^{3}$, Haomin Wen$^{4}$, Zhao Li$^{5}$, Meikang Qiu$^{6}$, Hongyan Li$^{1}$, \\ Ming Jin$^{2}$\thanks{Corresponding author}\And Shirui Pan$^{2*}$ 
\affiliations
$^1$Xidian University
$^2$Griffith University\\
$^3$Yunnan University
$^4$Carnegie Mellon University
$^5$Zhejiang University
$^6$Augusta University\\
$^7$The Hong Kong University of Science and Technology (Guangzhou)
\emails
yfge@stu.xidian.edu.cn,
jli226@connect.hkust-gz.edu.cn,
yjzhao@ynu.edu.cn,
haominwe@andrew.cmu.edu,
lzjoey@gmail.com,
qiumeikang@yahoo.com,\\
hyli@xidian.edu.cn,
mingjinedu@gmail.com,
s.pan@griffith.edu.au
}

\begin{document}

\maketitle

\begin{abstract}
Text-to-Time Series generation holds significant potential to address challenges such as data sparsity, imbalance, and limited availability of multimodal time series datasets across domains. While diffusion models have achieved remarkable success in Text-to-X (e.g., vision and audio data) generation, their use in time series generation remains in its nascent stages. Existing approaches face two critical limitations: (1) the lack of systematic exploration of general-proposed time series captions, which are often domain-specific and struggle with generalization; and (2) the inability to generate time series of arbitrary lengths, limiting their applicability to real-world scenarios.
In this work, we first categorize time series captions into three levels: point-level, fragment-level, and instance-level. Additionally, we introduce a new fragment-level dataset containing over 600,000 high-resolution time series-text pairs. Second, we propose \textbf{\texttt{\textcolor{Red}{T}}}ext-\textbf{\texttt{\textcolor{Green}{to}}}-\textbf{\texttt{\textcolor{Blue}{S}}}eries (\method), a diffusion-based framework that bridges the gap between natural language and time series in a domain-agnostic manner. \method employs a length-adaptive variational autoencoder to encode time series of varying lengths into consistent latent embeddings. On top of that, \method effectively aligns textual representations with latent embeddings by utilizing Flow Matching and employing Diffusion Transformer as the denoiser. We train \method in an interleaved paradigm across multiple lengths, allowing it to generate sequences of any desired length. Extensive evaluations demonstrate that \method achieves state-of-the-art performance across 13 datasets spanning 12 domains.
\end{abstract}
\vspace{-2mm}
\section{Introduction}
Time series generation (TSG) enables the creation of high-quality data in scenarios with limited available datasets, thus serving as a way to simulate diverse, multimodal temporal dynamics, which offers significant value in real-world applications.
Generating required objects from text (Text-to-X) is a research area that meets human needs and has potential at the present time. Driven by the success of diffusion models, Text-to-X generation has made remarkable strides in domains such as image generation~\cite{rombach2022high,esser2024scaling}, video generation~\cite{girdhar2023emu}, and speech processing~\cite{le2024voicebox}.
Specifically, diffusion models refine noisy data progressively through a learned process, ultimately producing high-fidelity outputs. While Text-to-X generation using diffusion models has been extensively explored in vision and audio data, their application to time series generation is still in its early stages.

Existing studies~\cite{yang2024survey} on TSG with diffusion models can be classified into three types based on the conditioning information. i) Label-based condition. A class-conditioned diffusion method for generating synthetic EEG signals is introduced in \cite{sharma2023medic}. ii) Temporal-based condition. Sensor data synthesis using statistical adjustments is explored in \cite{zuo2023unsupervised}. Although class- and temporal-conditioned generation is well discussed \cite{yuan2024diffusion,liu2024diffusion}, they are less flexible than text-based approaches \cite{gao2024lumina}. iii) Text-based condition. \cite{fu2024creating} uses domain-specific metadata (e.g., location, weather) to generate synthetic energy data, while \cite{wang2024multi} and \cite{lai2025diffusets,alcaraz2023diffusion} apply textual conditioning to synthesize sales and clinical ECG data, respectively. However, these methods are often domain-specific, their metadata captions can not support high resolution general alignment between time series and fine-grained captions.

Despite the progress made in applying diffusion models to time series modeling, two significant challenges remain in this domain. 
\textbf{First}, the scarcity of high-resolution general-propose text-time series caption datasets limits progress in Text-to-Time Series (T2S) generation, while existing datasets are domain-specific (e.g., healthcare \cite{johnson2023mimic}, economics \cite{cortis2017semeval}).
\textbf{Second}, existing TSG models \cite{yuan2024diffusion,desai2021timevae} typically require separate training for datasets of different lengths within each domain, making it challenging to develop length-arbitrary T2S models.
Most approaches rely on predefined sequence lengths tied to the training data, limiting their ability to generalize. Since real-world time series exhibit inherent variability in length due to factors such as data collection frequency or system-specific temporal dynamics, the need for length-specific training significantly hinders the scalability and practicality of these models.

To address these challenges, we introduce a new fragment-level dataset, \dataset, containing over 600,000 high-resolution fragment-level text-time series pairs, which serves as a foundation for exploring T2S generation. 
\method, a diffusion-based model is presented that bridges the gap between natural language and time series data in a domain-agnostic manner. 
Specifically, \method utilizes a length-adaptive variational autoencoder to encode time series of varying lengths into consistent latent embeddings. The model then aligns textual representations with latent embeddings using flow matching and employs a diffusion transformer as the denoiser. By training \method in an interleaved manner across diverse datasets, the model is able to generate high-quality and semantic aligned time series of arbitrary lengths during inference, overcoming the fixed-length limitations of prior approaches. All resources have been made available\footnote{\url{https://github.com/WinfredGe/T2S}}.
This work marks three key contributions: 
\begin{itemize}[left=0.2em]
    \item We systematically explore the existing T2S datasets, and introduce a novel, high-resolution fragment-level multimodal dataset for text-to-time series generation tasks. 
    \item We propose the first domain-agnostic model for text-to-time series generation, which integrates flow matching and the diffusion transformer, and is capable of generating semantically aligned time series of arbitrary lengths.
    \item \method sets a new state-of-the-art performance across 13 datasets from 12 domains in time series generation,  consistently outperforming both diffusion-based models and those based on large language models.
\end{itemize} 
\vspace{-2mm}
\section{Definition and Dataset}
\subsection{Problem Definition and Notation}
Let $ \mathbf{x} \in \mathbb{R}^L $  denote a univariate time series of length \( L \). Textual captions, represented as \( \mathbf{T} \), provide semantic guidance across varying levels of granularity.
\begin{definition}[Point-Level Description]
A point-level description \( \mathbf{T}_p \) provides semantic annotations for \textbf{individual time points} within the time series \( \mathbf{x} \). Each point-level description \( \mathbf{T}_p^{(j)} \) corresponds to the \( j \)-th time point, where \( j \in [1, \ldots, L] \), providing a fine-grained guidance for each point in time series \( \mathbf{x} \).
\end{definition}
\begin{definition}[Fragment-Level Description]
A fragment-level description \( \mathbf{T}_f \) provides semantic annotations for non-overlapping and \textbf{contiguous fragments} of the time series \( \mathbf{x} \). Each fragment-level description \( \mathbf{T}_f^{(j)} \) corresponds to the \( j \)-th fragment, where the length $|\mathbf{T}_f^{(j)}|$ is arbitrary and determined by the specific structure of time series \( \mathbf{x}\).
\end{definition}
\begin{definition}[Instance-Level Description]
An instance-level description \( \mathbf{T}_i \) provides a global semantic annotation for the \textbf{entire time series} \( \mathbf{x} \), providing high-level guidance encompassing all time points.
\end{definition}

\noindent We can use these captions to guide time series generation:
\begin{definition}[T2S Generation]
Given a text–time series dataset \( D = \{(\mathbf{x}^{(i)}, \mathbf{T}_{*}^{(i)})\}_{i=1}^{N} \) with \( N \) samples, the task of T2S generation aims to learn a generative model \( G \) that maximizes the conditional probability \( P_G(\mathbf{x}\mid\mathbf{T}_{*}) \), where \( \mathbf{T}_{*} \) is the textual guidance provided at one of the following levels (\( \mathbf{T}_p,\mathbf{T}_f, \mathbf{T}_i \)). The generated time series \( \mathbf{x} \) must semantically align with the textual guidance \( \mathbf{T}_{*} \).
\label{T2Sgeneration}
\end{definition}

\subsection{TSFragment-600K}
The proposed T2S generation task leverages textual captions at three granularity levels: point, fragment, and instance. While datasets for point-level \cite{liu2024time} and instance-level \cite{kawaguchi2024sushi} descriptions are readily available, fragment-level descriptions remain an underexplored area. Fragment-level descriptions strike a balance between the granularity of point-level annotations, which may overlook broader temporal patterns, and the holistic nature of instance-level descriptions, which might obscure local dependencies. By encapsulating local temporal trends while preserving meaningful contextual relationships, fragment-level descriptions provide an ideal framework for evaluating the generative capabilities of T2S models. 

\begin{figure}
    \centering
    \includegraphics[width=1\linewidth]{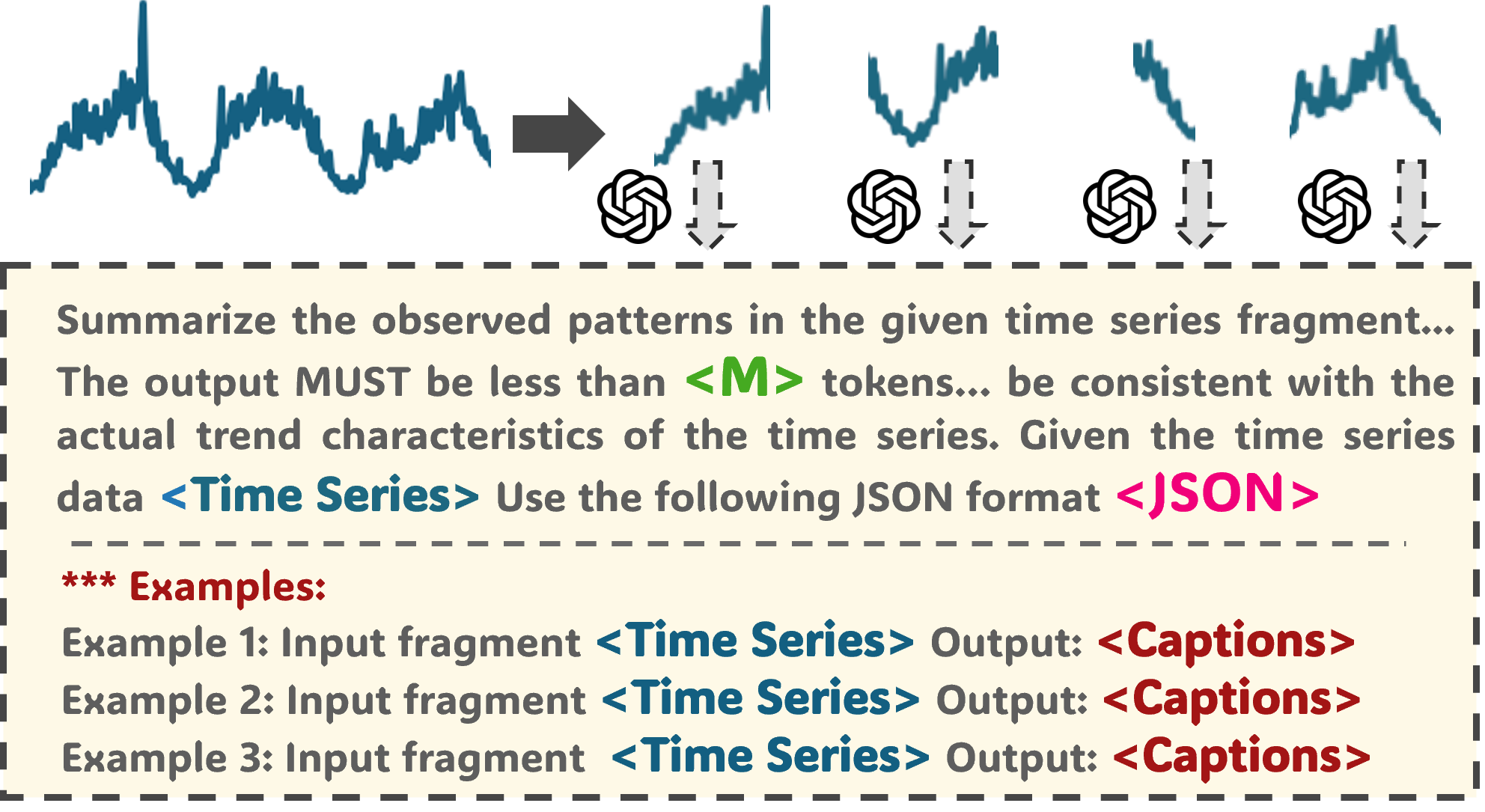}
    \caption{Dataset Generation. GPT-4o-mini to generate high-quality natural language descriptions for each time series fragment. }
    \vspace{-2mm}
    \label{fig:Dataset-Generation}
\end{figure}

To this end, we introduce \dataset, a novel dataset comprising over 600,000 fragment-level text-time series pairs. Each captions captures fine-grained temporal morphological characteristics, offering a rich and nuanced representation of the underlying trends. As illustrated in Figure~\ref{fig:Dataset-Generation}, we employ GPT-4o-mini to generate high-quality natural language descriptions for each time series fragment, focusing on local trends and variations. Unlike prior approaches that rely on predefined dictionaries of time-series changes \cite{imani2019putting}, our captions are expressed in natural language, enhancing their interpretability and applicability.

Specifically, we propose a novel generation pipeline to construct fragment-level captions for time series data.  
First, a univariate time series \( \mathbf{x} \) is segmented into $k$ non-overlapping fragments,with each fragment \( \mathbf{x}^{(j)} \in \mathbb{R}^{l_j} \) as a contiguous temporal segment for which textual captions are generated.
Second, a seed-based prompting strategy is designed to capture high-quality captions of fragments. To accurately capture temporal dynamics, human experts curate high-quality GPT-4o descriptions for a subset of fragments, which serve as representative seed prompts. These prompts guide GPT-4o-mini in generating concise, consistent, and semantically rich captions for all fragments. A token limit \textcolor[RGB]{69,170,33}{\(\langle \boldsymbol{M} \rangle\)} is applied to ensure a balance between informativeness and brevity.

Finally, five candidate captions are generated for each time series sample, and their embeddings are computed using \textsl{text-embedding-3-small}. We ensure the quality of generated textual captions by leveraging cosine similarity among their embeddings. Each caption is scored based on the average similarity of its embedding with others, and the one with the highest score is selected as the optimal text-time series pair, ensuring semantic alignment and coherence.
Using this pipeline, we generate reliable fragment-level descriptions for eight classical time series datasets across diverse domains, including energy consumption, financial, exchange rates, traffic, air quality, and meteorological variables. The resulting fragment-level dataset, \dataset, comprises over $600,000$ samples with corresponding captions.
\vspace{-1mm}
\section{Methodology}
\begin{figure*}[t]
	\centering 
	\includegraphics[width=1\textwidth]{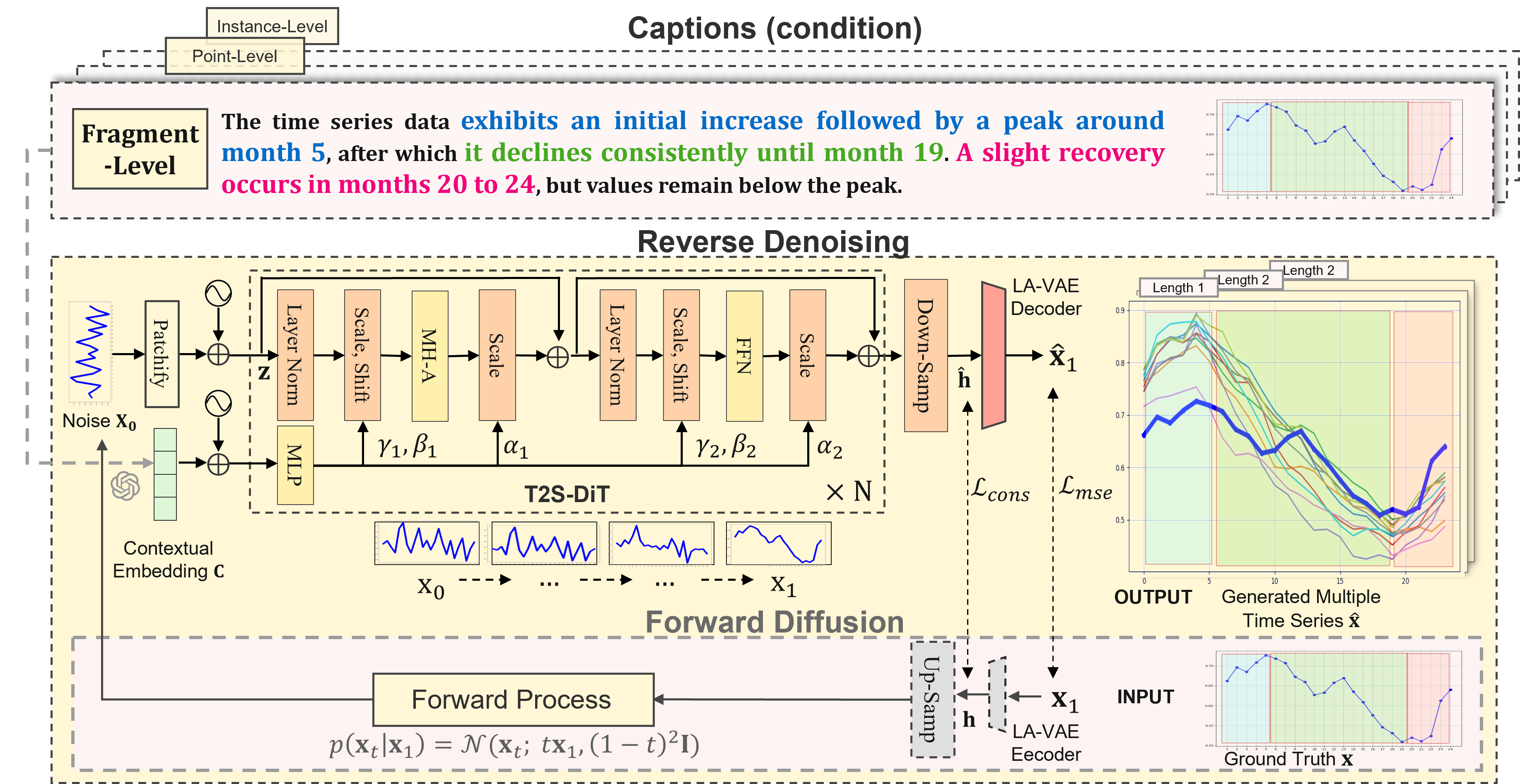}
	\caption{Overview of the \method model. The framework conditions on captions for time series generation. LA-VAE encodes variable-length inputs into a latent space and decodes outputs to the original length. Forward diffusion transforms the original time series into noise, while T2S-DiT performs reverse denoising to align textual and temporal features, generating high-quality time series. }
	\label{fig:architecture-overview} 
    \vspace{-2mm}
\end{figure*}

In this section, we introduce the \model architecture, as depicted in Figure \ref{fig:architecture-overview}. The architecture consists of two key components:
\begin{itemize}[left=-0.25em]
\item \textbf{T2S Diffusion Transformer (T2S-DiT)}: T2S-DiT facilitates high-resolution alignment between captions and the temporal latent space. It employs flow matching \cite{liu2022flow} as the diffusion backbone, the diffusion transformer module \cite{peebles2023scalable} as the denoiser. Within this denoiser, textual information is integrated with the input features through adaptive layer normalization

\item \textbf{Pretrained Length-Adaptive Variational Autoencoder (LA-VAE)}: LA-VAE encodes variable-length time series into a unified latent feature space and decodes them back to their original temporal dimensions. An interleaved training strategy is adopted to enable effective handling of varying input lengths during training.

\end{itemize}
We first employ LA-VAE to map time series of varying lengths into the latent space. The T2S-DiT module then denoises this latent space conditioned on the caption, aligning the textual and temporal features. During inference, a noise sequence of arbitrary length, encoded by LA-VAE, generates aligned time series based on the given caption.

\subsection{T2S Diffusion Model}

\textbf{Flow Matching Framework}. 
Inspired by~\cite{esser2024scaling,polyak2024movie} and experimental comparisons with DDPM, we adopt the flow matching framework~\cite{lipman2022flow}, specifically adopting the rectified flow approach~\cite{liu2022flow} using optimal transport paths. 
This framework offers superior generation quality and a more stable inference process compared to DDPM~\cite{ho2020denoising}, with reduced training costs.

Flow matching consists of forward and reverse processes. During training, given a time series sample in latent space \(\mathbf{z}_1\), a noisy sample \(\mathbf{z}_0 \sim \mathcal{N}(0,\mathbf{I})\), and a time step \(t \in [0, 1]\), the forward path \(\mathbf{z}_t\) is defined as:

\begin{equation}
    p(\mathbf{z}_t \mid \mathbf{z}_1) = \mathcal{N}\left(\mathbf{z}_t; t \mathbf{z}_1, (1-t)^2 \mathbf{I}\right),
    \label{eq:flow_matching_forward_prob}
\end{equation}
\noindent where $\mathbf{z}_t$ evolves along an optimal transport path defined by:

\begin{equation} 
\mathbf{z}_t = t \mathbf{z}_1 + (1-t) \mathbf{z}_0, \label{eq:flow_matching_forward} 
\end{equation}

\noindent and the ground truth velocity of this transition is described by: $\mathbf{v}_t = \frac{d\mathbf{z}_t}{dt} = \mathbf{z}_1 - \mathbf{z}_0$.
In the reverse process, the denoiser model aims to predict the velocity \( \mathbf{u}_{\theta}(\mathbf{z}_t, t, \mathbf{C}) \) , where $\mathbf{C}$ represents the text prompt embedding generated by \( \mathbf{T}_{*} \) in Definition \ref{T2Sgeneration}. The model minimizes the mean squared error (MSE) between the ground truth and predicted velocities:
\begin{equation} 
E_{\mathbf{z}_t, t, \mathbf{C}} \left| \mathbf{u}_{\theta}\left(\mathbf{z}_t, t, \mathbf{C} \right) - \mathbf{v}_t \right|^2. 
\end{equation}

During sampling, pure noise \( \mathbf{z}_0 \sim \mathcal{N}(0, 1) \) is iteratively denoised to obtain \( \mathbf{\hat{z}}_1 \) by solving the ordinary differential equation (ODE) with the well-trained denoiser \( \mathbf{u}_{\theta}\left(\mathbf{z}_t, t, \mathbf{C}\right) \).
\model adopts a classifier-free guidance framework, which does not rely on explicit class labels for conditioning. Unlike traditional class-based conditional generation methods, which use fixed condition categories to guide the generation process, classifier-free guidance provides a more flexible and effective way to balance unconditional and conditional generation.
During training, the method randomly sets the condition to zero by a random ratio. During inference, the model first performs conditional generation, followed by unconditional generation, and then combines the results using a guidance scale \(\delta\). The formula is:
\begin{equation}
\mathbf{u}_{\theta}\left(\mathbf{z}_t, t, \mathbf{C}\right) = (1+\delta) \mathbf{u}_{\theta}\left(\mathbf{z}_t, t, \mathbf{C}\right) - \delta \mathbf{u}_{\theta}\left(\mathbf{z}_t, t\right),
\end{equation}
\noindent where \( \mathbf{u}_{\theta}(\mathbf{z}_t, t, \mathbf{C}) \) is the noise estimater with input \( \mathbf{z}_t \) and condition \( \mathbf{C} \) and \( \mathbf{u}_{\theta}(\mathbf{z}_t, t) \) is the conditioned noise estimater .

\noindent\textbf{Diffusion Transformer}.
Building on the superior performance of the DiT in computer vision \cite{esser2024scaling} and recent advances in time series analysis \cite{chen2024visionts}, we developed a patchified DiT denoiser that leverages DiT's fine-grained visual feature extraction and its ability to capture subtle latent temporal patterns, aligning these patterns with corresponding textual descriptions through the 2D latent representations encoded via LA-VAE.               
The latent space representation of time series can be conceptualized as single-channel, two-dimensional grayscale images \( \mathbf{z}_t \). We first patchify \( \mathbf{z}_t \) into a sequence of tokens. Then the input tokens are obtained by summing the sequence of tokens with two-dimensional positional embeddings. \( \mathbf{c}_t\) and \( \mathbf{z}_t\) represent the conditioning associated with time step \(t\) and input tokens.
To achieve alignment between the conditioning \( \mathbf{c}_t \) and the input tokens \( \mathbf{z}_t \), we employ an adaptive layer normalization (AdaLN), a form of modulated layer normalization \cite{huang2017arbitrary}:

\begin{align}
&\operatorname{AdaLN}(\mathbf{z}_t, \mathbf{c}_t) = \gamma_t \left( \frac{\mathbf{z}_t - \mu_t}{\sigma_t} \right) + \beta_t,
\end{align}
\noindent where scaling parameter \( \alpha_{t,1}, \alpha_{t,2} ,\gamma_{t,1}, \gamma_{t,2} \) and shifting parameter \( \beta_{t,1}, \beta_{t,2} \) are chunked outputs, dynamically adjusted based on the textual information, as shown in Equation~\ref{func:conditioning_chunk}.
\begin{align}
\gamma_{t,1}, \gamma_{t,2}, \mu_{t,1}, \mu_{t,2}, \beta_{t,1}, \beta_{t,2} = \operatorname{MLP}(\mathbf{c}_t).
\label{func:conditioning_chunk}
\end{align}
The overall procedure can be formulated as follows:
\begin{equation}
     \mathbf{z}_t^{(1)} = \gamma_{t,1}\left(\frac{\mathbf{z}_t-\mu_t^{(1)}}{\sigma_t^{(1)}}\right)+\beta_{t,1}, 
     \mathbf{z}_t^{(2)} = \alpha_{t,1} \cdot \operatorname{MH-A}\left(\mathbf{z}_t^{(1)}\right), \\
\end{equation}
\begin{equation}
  \mathbf{z}_t^{(3)}    = \gamma_{t,2}\left(\frac{\mathbf{z}_t^{(2)}-\mu_t^{(2)}}{\sigma_t^{(2)}}\right)+\beta_{t,2}, 
    \mathbf{z}_t^{(4)} = \alpha_{t,2} \cdot \operatorname{FFN}\left(\mathbf{z}_t^{(3)}\right), \\
\end{equation}
\begin{equation}
    \mathbf{u}_{\theta}\left(\mathbf{z}_t, \mathbf{t}, \mathbf{c_t} \right) = \operatorname{MLP}\left(\operatorname{LayerNorm}\left(\frac{\mathbf{z}_t^{(4)} - \mu_t^{(4)})}{\sigma_t^{(4)}}\right)\right),
\end{equation}
\noindent where \(\mu_t^{(i)}\) and \(\sigma_t^{(i)}\) denote the mean and variance of the input \(\mathbf{z}_t\) for the \(i\)-th layer,  respectively, and $\operatorname{MH-A}$ represents the multi-head attention.
The entire process integrates adaptive layer norm into the transformer architecture, enabling it to dynamically adapt to textual conditioning. 
By leveraging this mechanism, the diffusion transformer aligns the input time series tokens with the contextual information, improving the generative performance of the model.

\subsection{Length-adaptive VAE}

T2S generation should support arbitrary-length generation to meet real-world application demands. For example, website user activity analysis and medical monitoring.
However, previous works \cite{lai2025diffusets} typically assume a fixed length, limiting their applicability.
To address this limitation, we propose a pretrained LA-VAE, enabling the modeling and generation of time series with arbitrary lengths. 
In this study, we mixed data with lengths of 24, 48, and 96, then trained them in a unified framework. During sampling, arbitrary-length data can be generated within a specified range.
Given an input time series \( \mathbf{x} \), the LA-VAE encoder transforms \( \mathbf{x} \) into a latent representation \(\mathbf{h}\). 
This latent representation \( \mathbf{h} \) is subsequently upsampled to a fixed-size latent embedding \(\mathbf{z} = \operatorname{Up-Samp}(\mathbf{h}_t)\), which serves as the input to the diffusion model. 
Through the diffusion process, a refined latent embedding \( \hat{\mathbf{z}}_t \) is generated. 
A downsampling operation is then applied to \( \hat{\mathbf{z}}_t \), yielding the latent vector \(\hat{\mathbf{h}} = \operatorname{Down-Samp}(\hat{\mathbf{z}}_t)\). 
Finally, the VAE decoder reconstructs \( \hat{\mathbf{h}} \) into a time series with the original length.
This process enables the handling of variable-length time series within a unified framework.

\noindent\textbf{Consistency loss}. Linear interpolation during upsampling and downsampling introduces blurriness and artifacts, as it fails to capture nonlinear features and signal curvature. To mitigate this, we introduce a latent space consistency loss term \(\operatorname{MSE}(\mathbf{h}, \hat{\mathbf{h}})\) to enhance reconstruction quality:
\begin{equation}
\mathcal{L}(\mathbf{x}, \hat{\mathbf{x}}, \mathbf{h}, \hat{\mathbf{h}}) = \operatorname{MSE}(\mathbf{x}, \hat{\mathbf{x}}) + \lambda \operatorname{MSE}(\mathbf{h}, \hat{\mathbf{h}}),
\label{eq:loss_sum}
\end{equation}
\noindent where the first term enforces fidelity to the original time series, and the second ensures consistency in the latent space.

\subsection{Interleaved Training}
Traditional sequential training paradigm often lead to catastrophic forgetting.
To effectively train models on datasets of varying lengths within a unified framework, we propose a novel interleaved training paradigm, outlined in Algorithm~\ref{alg:interleaved_learning}.

Assume a domain contains $d$ datasets with different lengths, denoted as $\{l_1, l_2, \dots, l_d\}$. During training, we shuffle all samples across these datasets and randomly sample $batch\_size$  samples for each batch. 
Within each iteration, training is interleaved across these batches, which improves the model's generalization ability.

\begin{algorithm}
\small
\caption{Interleaved Training for Mixed Datasets}
\label{alg:interleaved_learning}
\begin{algorithmic}[1]
\REQUIRE Datasets $\{D_1, \dots, D_k\}$, LA-VAE $\phi( \cdot )$, T2S-DiT $g_{\theta}( \cdot)$, batch\_size, number of iterations $N$.
\STATE $n_0 = 0$, $n_i \gets |D_i|$, $i = 1, \dots, k$
\STATE \texttt{DatasetSampling():}
\STATE \hspace{1em} $n \gets \sum_{i=1}^{k} n_i$
\STATE \hspace{1em} $j \sim \text{Uniform}(1, n)$
\STATE \hspace{1em} \text{Find } $m \in \mathbb{Z}^+$ \text{ such that } $j \in \left( \sum_{i=1}^{m} n_i, \sum_{i=1}^{m+1} n_i\right]$
\STATE \hspace{1em} \text{Return } $D_m[j - \sum_{i=1}^{m-1} n_i]$
\FOR{iter = 1 \textbf{~to~} $N$}
    \STATE $D \gets \{\}, \text{loss} = 0$
    \FOR{$i = 1 \textbf{~to~} \text{batch\_size}$}
        \STATE \hspace{1em} $D \gets D \cup \texttt{DatasetSampling()}$
    \ENDFOR
    \FOR{$\text{length } i = 1 \textbf{~to~} k$}
        \STATE $batch_i = \{ s \in D \mid \text{len}(s) = i \}$
        \STATE $\hat{\mathbf{S}} \gets g_{\theta}(\phi(batch_i))$
        \STATE \text{loss} = \text{loss} + $\mathcal{L}$ \text{ in equation }\ref{eq:loss_sum}
    \ENDFOR
    \STATE Update $\phi$
\ENDFOR   
\end{algorithmic}
\end{algorithm}
\vspace{-2mm}
\section{Experiments}
We conduct an extensive evaluation across 13 datasets spanning 12 domains to assess the performance of the \method, aiming to address the following key research questions:

\begin{itemize}[left=-0.08em]
    \item \textbf{RQ1}: How does \method\ compare in performance to existing state-of-the-art methods given fragment-level captions?
    \item \textbf{RQ2}: How does \method\ compare in performance to existing methods given point-level and instance-level captions?
    \item \textbf{RQ3}: How do the different components within the \method\  affect its overall generation performance?
    \item \textbf{RQ4}: How sensitive is the \method's performance to key hyperparameters, and does it require additional fine-tuning?
    \item \textbf{RQ5}: How effective is \method\ when trained on limited data?
\end{itemize}

\subsection{Experimental Settings}

\noindent\textbf{Datasets.} 
We evaluated our model on three distinct datasets: Point-Level, Fragment-Level, and Instance-Level.
\begin{itemize}[left=-0.08em]
\item \textbf{Point-Level Dataset}: 
The Time-MMD dataset \cite{liu2024time} links individual time series points with corresponding textual news, consisting of 23,618 data points across six domains, including Climate, Economy, and Social Goods. We adapted the dataset by concatenating each time series point with its associated text.
\item \textbf{Fragment-Level Dataset}: \dataset pairs time series data with captions across seven domain, including Electricity, ETT, Exchange, and Traffic \cite{wu2021autoformer}. It consists of over 600,000 samples.
\item \textbf{Instance-Level Dataset}: SUSHI, a simulated dataset \cite{kawaguchi2024sushi}, comprises 2,800 samples generated from 15 pre-defined functions. 

\end{itemize}

\begin{table*}[htbp]
  \centering
  \caption{Performance comparison on fragment level. All results are evaluated on three different metrics WAPE, MSE, and MRR@10. For interleaved training, arbitrary lengths of $\{24, 48, 96\}$ were selected, with evaluations performed separately for each length.}
  \resizebox{\textwidth}{!}{
  \Large
    \begin{tabular}{ccccc|ccc|ccc|ccc|ccc}
        \toprule
        & & \multicolumn{3}{c}{\method} & \multicolumn{3}{c}{DiffusionTS} & \multicolumn{3}{c}{TimeVAE} & \multicolumn{3}{c}{GPT-4o-mini} & \multicolumn{3}{c}{Llama3.1-8b} \\
        \cmidrule{3-17}
        Datasets & Length & \textcolor{someblue}{WAPE $\downarrow$} & \textcolor{somepurple}{MSE $\downarrow$} & \textcolor{someorange}{MRR@10 $\uparrow$} & \textcolor{someblue}{WAPE $\downarrow$} & \textcolor{somepurple}{MSE $\downarrow$} & \textcolor{someorange}{MRR@10 $\uparrow$} & \textcolor{someblue}{WAPE $\downarrow$} & \textcolor{somepurple}{MSE $\downarrow$} & \textcolor{someorange}{MRR@10 $\uparrow$} & \textcolor{someblue}{WAPE $\downarrow$} & \textcolor{somepurple}{MSE $\downarrow$} & \textcolor{someorange}{MRR@10 $\uparrow$} & \textcolor{someblue}{WAPE $\downarrow$} & \textcolor{somepurple}{MSE $\downarrow$} & \textcolor{someorange}{MRR@10 $\uparrow$} \\
      \midrule
      \multirow{3}[1]{*}{ETTh1} & 24   & \textbf{0.183} & \textbf{0.008} & \textbf{0.283} & 0.793 & 0.077 & 0.267 & 0.666 & 0.055 & 0.211 & 0.264 & 0.041 & 0.104 & 0.883 & 0.663 & 0.097 \\
      & 48    & \textbf{0.234} & \textbf{0.013} & 0.289 & 1.207 & 0.120 & \textbf{0.298} & 0.647 & 0.055 & 0.286 & 0.414 & 0.080 & 0.100 & 0.923 & 1.260 & 0.086 \\
      & 96    & \textbf{0.229} & \textbf{0.011} & \textbf{0.291} & 0.498 & 0.028 & 0.214 & 0.643 & 0.055 & 0.286 & 0.500 & 0.118 & 0.096 & 0.949 & 1.748 & 0.056 \\
      \midrule
      \multirow{3}[1]{*}{ETTm1}& 24    & 0.426 & 0.033 & \textbf{0.286} & 0.604 & 0.040 & 0.251 & 0.666 & 0.048 & 0.219 & \textbf{0.244} & \textbf{0.031} & 0.101 & 1.134 & 0.798 & 0.099 \\
      & 48    & 0.53 & \textbf{0.053} & \textbf{0.283} & 1.119 & 0.100 & 0.285 & 0.636 & 0.051 & 0.217 & \textbf{0.453} & 0.112 & 0.097 & 1.074 & 1.496 & 0.079 \\
      & 96    & \textbf{0.414} & 0.041 & \textbf{0.299} & 0.546 & \textbf{0.031} & 0.293 & 0.664 & 0.057 & 0.208 & 0.706 & 0.395 & 0.091 & 1.079 & 1.761 & 0.057 \\
      \midrule
      \multirow{3}[1]{*}{Electricity} & 24    & \textbf{0.135} & \textbf{0.010} & \textbf{0.28} & 0.617 & 0.041 & 0.253 & 0.207 & 0.016 & 0.213 & 0.734 & 0.592 & 0.092 & 0.926 & 1.140 & 0.064 \\
      & 48 & \textbf{0.155} & \textbf{0.013} & \textbf{0.244} & 1.128 & 0.102 & 0.227 & 0.208 & 0.017 & 0.216 & 1.014 & 1.065 & 0.068 & 1.038 & 1.416 & 0.054 \\
      & 96 & 0.238 & 0.031 & \textbf{0.318} & 0.545 & 0.032 & 0.247 & \textbf{0.213} & \textbf{0.018} & 0.257 & 1.024 & 1.210 & 0.059 & 1.085 & 1.740 & 0.034 \\
      \midrule
      \multirow{3}[1]{*}{Exchange Rate} & 24    & \textbf{0.292} & \textbf{0.033} & \textbf{0.334} & 0.791 & 0.077 & 0.272 & 1.165 & 0.105 & 0.252 & 1.072 & 2.060 & 0.052 & 1.258 & 2.052 & 0.045 \\
      & 48 & \textbf{0.259} & \textbf{0.033} & \textbf{0.315} & 1.217 & 0.122 & 0.298 & 1.064 & 0.106 & 0.306 & 0.933 & 1.074 & 0.082 & 1.562 & 2.125 & 0.051 \\
      & 96 & \textbf{0.48} & \textbf{0.047} & \textbf{0.31} & 0.504 & 0.048 & 0.216 & 0.977 & 0.106 & 0.274 & 1.141 & 1.625 & 0.054 & 1.433 & 1.892 & 0.055 \\
      \midrule
      \multirow{3}[1]{*}{Air Quality} & 24    & 0.884 & \textbf{0.02} & \textbf{0.304} & 0.806 & 0.078 & 0.265 & 2.303 & 0.022 & 0.302 & \textbf{0.557} & 0.379 & 0.093 & 0.878 & 0.697 & 0.085 \\
      & 48 & 1.295 & \textbf{0.044} & \textbf{0.297} & 1.439 & 0.120 & 0.221 & 1.648 & 0.023 & 0.271 & \textbf{0.791} & 0.715 & 0.08 & 1.141 & 1.642 & 0.046 \\
      & 96 & 1.377 & 0.049 & \textbf{0.34} & \textbf{0.508} & 0.028 & 0.304 & 1.270 & \textbf{0.024} & 0.301 & 0.928 & 1.127 & 0.061 & 1.085 & 1.551 & 0.050 \\
      \midrule
      \multirow{3}[1]{*}{Traffic} & 24 & \textbf{0.353} & \textbf{0.005} & 0.201 & 0.795 & 0.077 & 0.220 & 0.544 & 0.008 & \textbf{0.233} & 1.260 & 1.912 & 0.020 & 1.144 & 1.938 & 0.022 \\
      & 48 & \textbf{0.506} & \textbf{0.008} & \textbf{0.219} & 1.202 & 0.120 & 0.188 & 0.594 & 0.011 & 0.211 & 1.189 & 1.928 & 0.011 & 1.138 & 1.988 & 0.004 \\
      & 96 & \textbf{0.543} & \textbf{0.01} & \textbf{0.262} & 0.509 & 0.028 & 0.171 & 0.641 & 0.013 & 0.207 & 1.18 & 2.093 & 0.010 & 1.107 & 1.994 & 0.001 \\
      \midrule
      \multicolumn{1}{c}{$1^{\text{st}}$ count} & -  & 12 & 14 & 16 & 1 & 1 & 1 & 1 & 2 & 1 & 4 & 1 & 0 & 0 & 0 & 0 \\
        \bottomrule
    \end{tabular}
}
\begin{tablenotes}    
\scriptsize          
\item * \model's interleaved training strategy enables cross-length training within each dataset, whereas other full-trained models require training and evaluation for fixed-length inputs.   
\end{tablenotes}       
  \label{tab:main_results}
\end{table*}

\noindent\textbf{Evaluation Metrics.}
We evaluated performance using Mean Squared Error (MSE), Weighted Absolute Percentage Error (WAPE) \cite{shao2024exploring}, and Mean Reciprocal Rank at 10 (MRR@10)\cite{craswell2009mean}.
\begin{itemize}[left=-0.08em]
    \item Weighted Absolute Percentage Error (WAPE):
    \begin{equation}\operatorname{WAPE}(y, \hat{y})=\frac{\sum_{i \in \Omega}\left|y_i-\hat{y}_i\right|}{\sum_{i \in \Omega}\left|y_i\right|},
    \end{equation}
    \noindent where $\Omega$ and $\hat{\Omega}$ represent truth space and generative space, respectively, while $y_i$ and $\hat{y}_i$ denote the corresponding $i$-th sample. WAPE is scale-invariant, making it suitable for time series reconstruction.

    \item Mean Reciprocal Rank at 10 (MRR@10):
    \begin{equation}
        \operatorname{MRR@10} = \frac{1}{|\Omega|} \sum_{i \in \Omega} \frac{1}{\text{rank}_i},
    \end{equation}
    \noindent where \(\text{rank}_i=\text{argmin}(n\mid\cos(\hat{y}_{i,n}, y_i)>\text{threshold})\).
    Here \(\text{rank}_i\) denotes the rank of the first relevant result for sample \(i\) among 10 results. The operator \(\text{argmin}(n \mid \cdots)\) returns the index of the first relevant result that satisfies the condition based on the cosine similarity \(\cos(\hat{y}_{i,n}, y_i)\) between the generated results and the truth \cite{ito2024clasp}.
\end{itemize}

\noindent\textbf{Baselines.}
To evaluate \model, we compared its performance against state-of-the-art fully trained models and zero-shot large language models, encompassing diverse paradigms to robustly assess its effectiveness. Among fully trained models, DiffusionTS \cite{yuan2024diffusion} uses diffusion with text and time embeddings for time-series generation and adapts to text-conditional generation by injecting text embeddings into its encoder and decoder using AdaLayerNorm. Meanwhile, TimeVAE \cite{desai2021timevae} utilizes a variational autoencoder framework with caption embeddings, using dense layers with ReLU to fuse input and text for conditioning. For the zero-shot baselines, GPT-4o \cite{openai2023gpt4omini} and Llama-3.1-8b \cite{dubey2024llama} are employed. To curb hallucination and ensure reliable outputs, Llama‑3 is guided by domain‑specific prompts, a repeat generation loop, and targeted post‑processing.

\begin{table*}[t]
    \centering
    \caption{Performance comparison on point and instance levels.  All results are evaluated on three different metrics WAPE, MSE, and MRR@10. For the instance-level dataset, SUSHI is used for training and inference with a fixed length of 2048. For point-level training, arbitrary lengths of \{24, 48, 96\} were selected for interleaved training.}
    \resizebox{\textwidth}{!}{ 
    \Large
    \begin{tabular}{ccccc|ccc|ccc|ccc|ccc}
        \toprule
        & & \multicolumn{3}{c}{\method} & \multicolumn{3}{c}{DiffusionTS} & \multicolumn{3}{c}{TimeVAE} & \multicolumn{3}{c}{GPT-4o-mini} & \multicolumn{3}{c}{Llama3.1-8b} \\
        \cmidrule{3-17}
        Datasets & Length & \textcolor{someblue}{WAPE $\downarrow$} & \textcolor{somepurple}{MSE $\downarrow$} & \textcolor{someorange}{MRR@10 $\uparrow$} & \textcolor{someblue}{WAPE $\downarrow$} & \textcolor{somepurple}{MSE $\downarrow$} & \textcolor{someorange}{MRR@10 $\uparrow$} & \textcolor{someblue}{WAPE $\downarrow$} & \textcolor{somepurple}{MSE $\downarrow$} & \textcolor{someorange}{MRR@10 $\uparrow$} & \textcolor{someblue}{WAPE $\downarrow$} & \textcolor{somepurple}{MSE $\downarrow$} & \textcolor{someorange}{MRR@10 $\uparrow$} & \textcolor{someblue}{WAPE $\downarrow$} & \textcolor{somepurple}{MSE $\downarrow$} & \textcolor{someorange}{MRR@10 $\uparrow$} \\
      \midrule
        \multirow{1}[1]{*}{SUSHI} & 2048 & 0.494 & 0.088 & \textbf{0.314} & \textbf{0.407} & \textbf{0.032} & 0.269 & 0.445 & 0.061 & 0.288 & 1.093 & 0.990 & 0.058 & 0.869 & 0.827 & 0.055 \\
        \midrule

        \multirow{3}[1]{*}{Agriculture} & 24 & \textbf{0.183} & \textbf{0.013} & \textbf{0.661} & 0.648 & 0.046 & 0.294 & 1.309 & 0.087 & 0.251 & 1.284 & 2.190 & 0.070 & 0.648 & 0.402 & 0.124 \\

        & 48 & \textbf{0.197} & \textbf{0.008} & 0.209 & 1.422 & 2.306 & 0.256 & 1.165 & 0.106 & \textbf{0.279} & 1.321 & 2.146 & 0.071 & 1.012 & 0.703 & 0.104 \\

        & 96 & \textbf{0.124} & \textbf{0.014} & \textbf{0.619} & 1.073 & 0.096 & 0.319 & 0.930 & 0.076 & 0.291 & 1.422 & 2.306 & 0.052 & 1.283 & 1.125 & 0.114 \\
        \midrule

        \multirow{3}[1]{*}{Climate} & 24 & \textbf{0.328} & \textbf{0.016} & \textbf{0.405} & 0.791 & 0.068 & 0.293 & 0.575 & 0.054 & 0.306 & 1.038 & 0.800 & 0.104 & 1.207 & 1.800 & 0.053 \\
        
        & 48 & \textbf{0.211} & \textbf{0.007} & \textbf{0.39} & 0.554 & 0.037 & 0.305 & 0.513 & 0.051 & 0.335 & 1.014 & 0.997 & 0.092 & 1.284 & 2.124 & 0.046 \\
        
        & 96 & \textbf{0.294} & \textbf{0.021} & \textbf{0.476} & 1.279 & 0.203 & 0.264 & 0.494 & 0.057 & 0.275 & 1.057 & 1.335 & 0.069 & 1.167 & 1.870 & 0.049 \\
        \midrule

        \multirow{3}[1]{*}{Economy} & 24 & \textbf{0.118} & \textbf{0.010} & \textbf{0.561} & 0.989 & 0.086 & 0.292 & 0.476 & 0.084 & 0.316 & 0.295 & 0.071 & 0.130 & 1.194 & 1.690 & 0.059 \\

        & 48 & \textbf{0.071} & \textbf{0.004} & \textbf{0.488} & 1.239 & 0.132 & 0.290 & 0.607 & 0.129 & 0.314 & 0.339 & 0.063 & 0.112 & 0.501 & 0.270 & 0.096 \\

        & 96 & \textbf{0.113} & \textbf{0.01} & \textbf{0.667} & 0.826 & 0.083 & 0.293 & 0.597 & 0.110 & 0.329 & 0.539 & 0.198 & 0.124 & 0.615 & 0.321 & 0.100 \\
        \midrule

        \multirow{3}[1]{*}{Energy} & 24 & \textbf{0.212} & \textbf{0.005} & \textbf{0.391} & 0.452 & 0.028 & 0.309 & 2.287 & 0.083 & 0.281 & 1.327 & 1.952 & 0.058 & 1.807 & 1.897 & 0.068 \\

        & 48 & \textbf{0.174} & \textbf{0.003} & \textbf{0.452} & 0.373 & 0.031 & 0.295 & 2.012 & 0.086 & 0.268 & 1.408 & 1.949 & 0.056 & 1.139 & 1.935 & 0.065 \\

        & 96 & \textbf{0.372} & \textbf{0.017} & \textbf{0.450} & 0.391 & 0.030 & 0.290 & 1.716 & 0.099 & 0.290 & 1.256 & 1.904 & 0.043 & 1.097 & 1.593 & 0.068 \\
        \midrule

        \multirow{3}[1]{*}{Health US} & 24 & \textbf{0.328} & \textbf{0.009} & 0.192 & 0.427 & 0.048 & \textbf{0.224} & 0.888 & 0.051 & 0.323 & 1.008 & 1.789 & 0.050 & 1.230 & 1.982 & 0.068 \\

        & 48 & \textbf{0.264} & \textbf{0.008} & 0.129 & 0.424 & 0.052 & \textbf{0.221} & 0.743 & 0.051 & 0.296 & 1.045 & 1.930 & 0.014 & 1.163 & 1.883 & 0.035 \\

        & 96 & \textbf{0.316} & \textbf{0.012} & 0.141 & 0.594 & 0.073 & \textbf{0.215} & 0.753 & 0.051 & 0.308 & 1.089 & 1.940 & 0.002 & 1.176 & 1.957 & 0.013 \\
        \midrule

        \multirow{3}[1]{*}{Social Goods} & 24 & 0.901 & \textbf{0.024} & \textbf{0.583} & \textbf{0.640} & 0.070 & 0.305 & 0.942 & 0.049 & 0.327 & 1.789 & 1.420 & 0.093 & 1.353 & 1.653 & 0.058 \\

        & 48 & 0.721 & 0.082 & \textbf{0.452} & \textbf{0.410} & \textbf{0.045} & 0.337 & 0.678 & 0.049 & 0.349 & 1.390 & 1.920 & 0.055 & 1.247 & 1.862 & 0.056 \\

        & 96 & \textbf{0.283} & \textbf{0.020} & \textbf{0.494} & 0.417 & 0.041 & 0.310 & 0.677 & 0.054 & 0.260 & 1.347 & 1.634 & 0.077 & 1.261 & 1.670 & 0.030 \\
        \midrule

        \multicolumn{1}{c}{$1^{\text{st}}$ count} & - & 16 & 17 & 15 & 3 & 2 & 3 & 0 & 0 & 1 & 0 & 0 & 0 & 0 & 0 & 0 \\
        \bottomrule
    \end{tabular}
    } 
  \begin{tablenotes}    
    \scriptsize          
    \item * \model's interleaved training strategy enables cross-length training within each dataset, whereas other full-trained models require training and evaluation for fixed-length inputs.   
  \end{tablenotes}       
    \label{tab:auxiliary_results}
\end{table*}

\subsection{Performance Comparison on Fragment-Level Descriptions (RQ1)}

Table~\ref{tab:main_results} presents the fragment-level performance comparison across six datasets, evaluated using WAPE, MSE, and MRR@10. \model achieves top performance across all metrics, securing 14 out of 18 entries (77.8\%) for MSE, significantly outperforming DiffusionTS (5.6\%) and TimeVAE (11.1\%). 
On the exchange rate dataset, \model\ achieves an average MSE of 0.039, representing a 56.0\% improvement over DiffusionTS and a 68.9\% improvement over TimeVAE.
These results demonstrate \model's superior ability to align textual and temporal features across diverse datasets and fragment lengths.
Moreover, \model's interleaved training strategy enables cross-length training within each dataset, removing the need for length-specific training required by baseline models, thereby enhancing scalability and generalization.

\subsection{Performance Comparison on Point and Instance-Level Descriptions (RQ2)}
Table~\ref{tab:auxiliary_results} presents the performance comparison at the point and instance levels across seven datasets. These evaluations assess models' abilities to capture fine-grained temporal annotations and generate coherent global patterns.
T2S outperforms all baselines across three metrics. 
It consistently achieves the lowest MSE values in 17 out of 18 entries, surpassing DiffusionTS and TimeVAE, which struggle to capture fine-grained temporal variations. Similarly, T2S secures the best WAPE scores in 16 out of 18 entries, demonstrating its robustness.
T2S achieves the top MRR@10 score of 0.314 on the instance-level SUSHI dataset, its WAPE and MSE results show an advantage over zero-shot models, only a slight underperformance compared to DiffusionTS in certain scenarios.
In summary, T2S demonstrates state-of-the-art performance by effectively balancing fine-grained precision with high-level semantic understanding.

\subsection{Ablation Study (RQ3)}

\begin{figure}[h]
    \centering
    \includegraphics[width=1\linewidth]{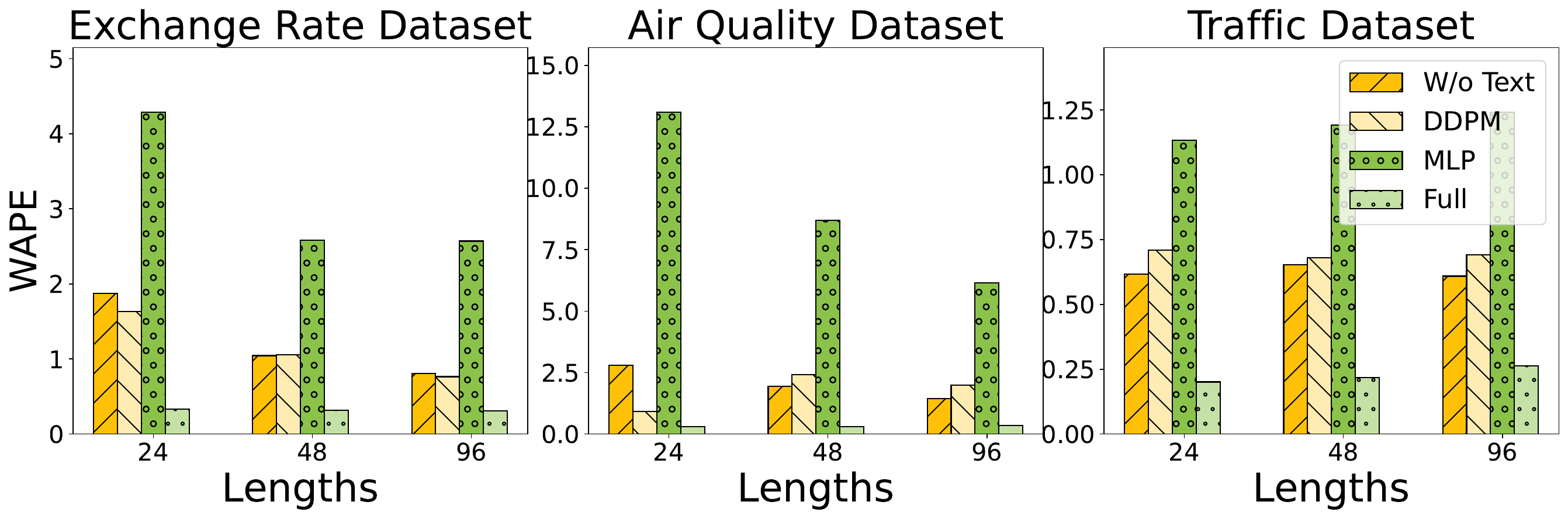}
    \caption{WAPE comparison across three datasets and generation lengths \{24, 48, 96\}. The legend indicates model configurations: W/o Text (no text guidance), DDPM (baseline with diffusion probabilistic modeling), MLP (denoiser replaced with Multi-Layer Perception), and Full model.}
    \vspace{-2mm}
    \label{fig:Heatmap}
\end{figure}

As shown in Figure \ref{fig:Heatmap} ,we conducted three ablation experiments on three datasets: Exchange Rate, Air Quality, and Traffic, evaluating nine  different settings. 
First, replacing the flow matching backbone with DDPM resulted in consistent performance drops, with an average error increase of 311.00\% across all datasets. 
Second, substituting the DiT denoiser with an MLP drastically degraded results, with errors rising by 877.67\% on the Exchange Rate dataset. Lastly, removing text guidance severely impacted high-resolution generation, with average error increase of 495.13\%, 327.10\%, 205.23\% across datasets with lengths \{24, 48, 96\}, respectively, highlighting the critical role of text in guiding generation quality.
These findings clearly highlight the critical role of each component, demonstrating that the full model is essential for achieving high-quality generation.

\begin{figure}[h]
    \centering
    \includegraphics[width=1\linewidth]{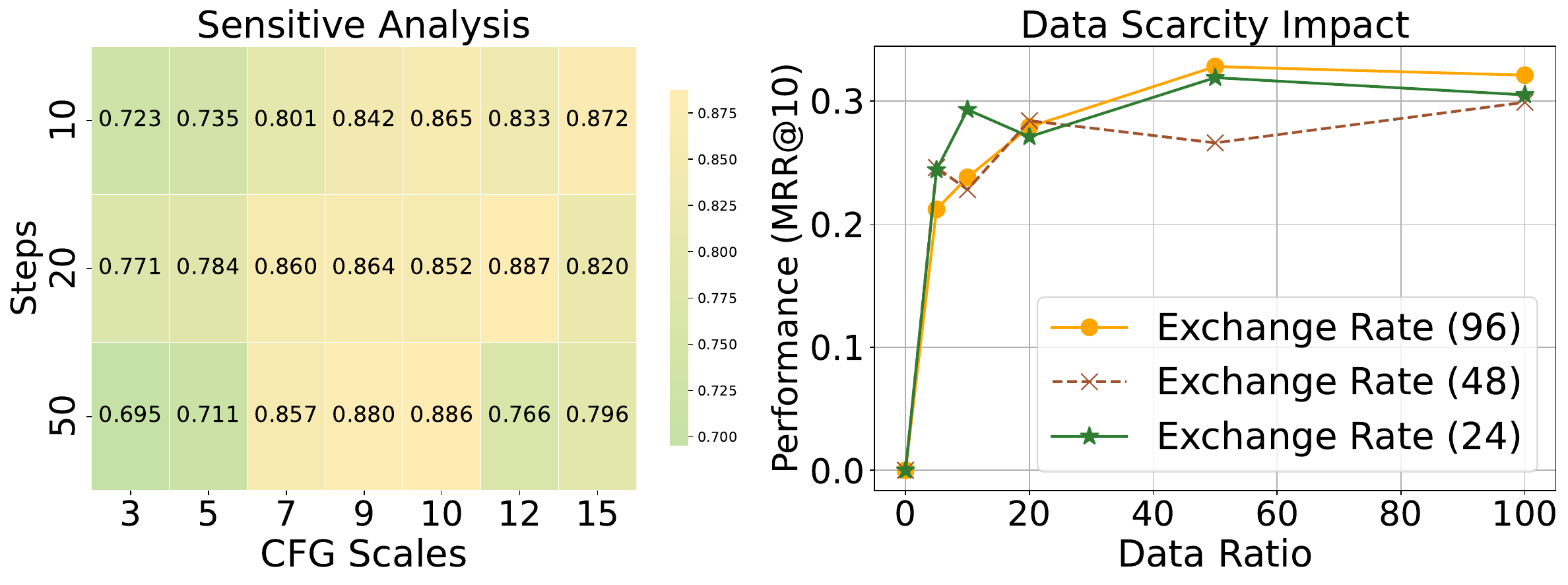}
    \caption{Parameter sensitivity analysis on the Exchange Rate dataset using MRR@10 (left). Data sparsity performance of \model on different data ratios. Consistent improvements are reported with an increasing data ratio (right).}
    \vspace{-2mm}
    \label{fig:sensitivity}
\end{figure}

\subsection{Parameter Sensitivity (RQ4)}
Recent study \cite{li2024learning} demonstrates the pivotal role of the inference stage in affecting the performance of diffusion models. Building on this, we explored the sensitivity of the flow matching diffusion model to key inference parameters: classifier-free guidance scales (CFG) and generation time steps,evaluated using MRR@10.
Figure ~\ref{fig:sensitivity} shows a heatmap illustrating performance impact, with yellow regions yielding superior results and green areas reflecting suboptimal performance. Notably, the model achieves higher MRR@10 scores within the range of CFG scores between 7 and 10 and generation time steps between 20 and 50. 
This analysis underscores the importance of precise inference-stage parameter selection in optimizing the performance of Flow Matching models.

\subsection{Data Scarcity (RQ5)}
To explore the impact of dataset size on the model’s performance, we evaluate the model on the Exchange Rate dataset under varying dataset scales. Specifically, we train and evaluate the model on subsets consisting of different proportions of the full dataset.
As shown in figure~\ref{fig:sensitivity}, the model demonstrates consistent improvement with increasing dataset size. Notably, using only 50\% of the dataset, the model achieves 93.8\% of the full dataset performance. Similarly, for the 48- and 96-length generations, the model reaches 92.4\% and 91.3\% of the full dataset performance, respectively. These results highlight the model’s strong generalization ability in data-scarce scenarios, showing its capacity to generate high-quality time-series even with limited training data.
\vspace{-1mm}
\section{Related Work}
\textbf{Text-Time Series Datasets}.
Existing text-time series pair datasets can be categorized into three types: instance-level, point-level, and fragment-level, based on how the caption and time series are temporally aligned.
At the point level, each time series point is paired with an event description, such as financial news or clinical notes \cite{yu2023temporal,cortis2017semeval,liu2024time}.
At the instance level, \texttt{TRUCE} \cite{jhamtani2021truth} and \texttt{SUSHI} \cite{kawaguchi2024sushi} utilize time series features, such as upward trends and peaks, as dictionary entries to generate coherent signals.
At the fragment level, several researchers have introduced datasets tailored for time series reasoning tasks \cite{williams2024context,chow2024towards}. 
A large-scale, fine-grained, general-purpose text-time series dataset for time series generation tasks remains in its early exploration.

\noindent\textbf{Text-Time Series Generation}.
The general text-to-time series paradigm can be achieved through contrastive learning or generative modeling. 
Recently, contrastive learning has been employed to facilitate text-to-time series mapping. However, these approaches are primarily focused on retrieval tasks \cite{ito2024clasp,rizhko2024astrom} and cannot be directly applied to time series generation.
In contrast, generative modeling, including variational autoencoders (VAEs) \cite{desai2021timevae,lee2023vector}, diffusion models \cite{yuan2024diffusion,kong2020diffwave,wen2023diffstg,narasimhan2024time}, and large language models \cite{openai2023gpt4omini,dubey2024llama}, provides more versatile frameworks for generating time series conditioned on textual descriptions. 
Among these, conditional diffusion models \cite{yuan2024diffusion,cao2024timedit,narasimhan2024time} show promise for text-to-time series generation due to their ability to model complex temporal dynamics and generate temporally coherent sequences.
For instance, time series generation conditioned on healthcare metadata \cite{alcaraz2023diffusion,lai2025diffusets} and sensor metadata \cite{zuo2023unsupervised,fu2024creating} has been explored. However, these methods are often domain-specific and fail to address the more general alignment between time series and their corresponding captions, limiting their broader applicability.
\vspace{-3mm}
\section{Conclusion}
We proposed \dataset, a high-resolution fragment-level multimodal dataset for text-to-time series generation tasks, and \model, the first domain-agnostic model for general text-to-time series generation. Leveraging LA-VAE and T2S-DiT, \model generates semantically aligned time series of arbitrary lengths with high fidelity. Comprehensive validation across 12 diverse domains demonstrates \method's superior performance, establishing a robust foundation for text-to-time series generation. 
\bibliographystyle{named}
\bibliography{ijcai25}

\end{document}